\documentclass{article} % For LaTeX2e
\usepackage{iclr2017_conference,times}

\usepackage{url}            % simple URL typesetting
\usepackage{booktabs}       % professional-quality tables
\usepackage{amsfonts,amssymb,amsmath}       % blackboard math symbols
\usepackage{nicefrac}       % compact symbols for 1/2, etc.
\usepackage{color}
\usepackage{graphicx}
\usepackage{multibib}
\usepackage{enumitem}
\usepackage{algorithm}
\usepackage{algpseudocode}

\newcites{appendix}{References}
\setlist[itemize]{leftmargin=*}

\title{Metacontrol for Adaptive Imagination-Based Optimization}

\iclrfinalcopy

\author{%
  Jessica B.~Hamrick\\
  UC Berkeley \& DeepMind\\
  \texttt{jhamrick@berkeley.edu} \\
  \And
  Andrew J.~Ballard \\
  DeepMind \\
  \texttt{aybd@google.com} \\
  \And
  Razvan Pascanu \\
  DeepMind \\
  \texttt{razp@google.com} \\
  \AND
  Oriol Vinyals \\
  DeepMind \\
  \texttt{vinyals@google.com} \\
  \And
  Nicolas Heess \\
  DeepMind \\
  \texttt{heess@google.com} \\
  \And
  Peter W.~Battaglia \\
  DeepMind \\
  \texttt{peterbattaglia@google.com} \\
}

\DeclareMathOperator*{\argmin}{arg\,min}

\newcommand{\controller}[0]{\pi^C}
\newcommand{\manager}[0]{\pi^M}

\begin{document}

\maketitle

\begin{abstract}
Many machine learning systems are built to solve the hardest examples of a particular task, which often makes them large and expensive to run---especially with respect to the easier examples, which might require much less computation. For an agent with a limited computational budget, this ``one-size-fits-all'' approach may result in the agent wasting valuable computation on easy examples, while not spending enough on hard examples. Rather than learning a single, fixed policy for solving all instances of a task, we introduce a \emph{metacontroller} which learns to optimize a sequence of ``imagined'' internal simulations over predictive models of the world in order to construct a more informed, and more economical, solution. The metacontroller component is a model-free reinforcement learning agent, which decides both \emph{how many} iterations of the optimization procedure to run, as well as \emph{which} model to consult on each iteration. The models (which we call ``experts'') can be state transition models, action-value functions, or any other mechanism that provides information useful for solving the task, and can be learned on-policy or off-policy in parallel with the metacontroller. When the metacontroller, controller, and experts were trained with ``interaction networks'' \citep{Battaglia2016} as expert models, our approach was able to solve a challenging decision-making problem under complex non-linear dynamics. The metacontroller learned to adapt the amount of computation it performed to the difficulty of the task, and learned how to choose which experts to consult by factoring in both their reliability and individual computational resource costs. This allowed the metacontroller to achieve a lower overall cost (task loss plus computational cost) than more traditional fixed policy approaches. These results demonstrate that our approach is a powerful framework for using rich forward models for efficient model-based reinforcement learning.
\end{abstract}

\section{Introduction}

%% describe the problem
While there have been significant recent advances in deep reinforcement learning \citep{Mnih2015,Silver2016} and control \citep{Lillicrap2015,Levine2016}, most efforts train a network that performs a fixed sequence of computations.
Here we introduce an alternative in which an agent uses a \emph{metacontroller} to choose which, and how many, computations to perform.
It ``imagines'' the consequences of potential actions proposed by an actor module, and refines them internally, before executing them in the world. The metacontroller adaptively decides which expert models to use to evaluate candidate actions, and when it is time to stop imagining and act. The learned experts may be state transition models, action-value functions, or any other function that is relevant to the task, and can vary in their accuracy and computational costs. Our metacontroller's learned policy can exploit the diversity of its pool of experts by trading off between their costs and reliability, allowing it to automatically identify which expert is most worthwhile.

We draw inspiration from research in cognitive science and neuroscience which has studied how people use a meta-level of reasoning in order to control the use of their internal models and allocation of their computational resources. Evidence suggests that humans rely on rich generative models of the world for planning \citep{Glaescher2010}, control \citep{Wolpert1998}, and reasoning \citep{Hegarty2004,johnson2010mental,Battaglia2013}, that they adapt the amount of computation they perform with their model to the demands of the task \citep{Hamrick2015}, and that they trade off between multiple strategies of varying quality \citep{Lee2014,Lieder2014,Lieder2016,Koolinpress}.

Our imagination-based optimization approach is related to classic artificial intelligence research on bounded-rational \emph{metareasoning} \citep{Horvitz1988,Russel1991,Hay2012}, which formulates a meta-level MDP for selecting computations to perform, where the computations have a known cost.
We also build on classic work by \cite{schmidhuber1990line,schmidhuber1990reinforcement}, which used an RL controller with a recurrent neural network (RNN) world model to evaluate and improve upon candidate controls online.

Recently \citet{Andrychowicz2016} used a fully differentiable deep network to learn to perform gradient descent optimization, and \citet{Tamar2016} used a convolutional neural network for performing value iteration online in a deep learning setting.
In other similar work, \citet{Fragkiadaki2015} made use of ``visual imaginations'' for action planning.
Our work is also related to recent notions of ``conditional computation'' \citep{Bengio2013,Bengio2015}, which adaptively modifies network structure online, and ``adaptive computation time'' \citep{Graves2016} which allows for variable numbers of  internal ``pondering'' iterations to optimize computational cost.

Our work's key contribution is a framework for \emph{learning to optimize} via a metacontroller which manages an adaptive, imagination-based optimization loop. This represents a hybrid RL system where a model-free metacontroller constructs its decisions using an actor policy to manage model-free and model-based experts.
Our experimental results demonstrate that a metacontroller can flexibly allocate its computational resources on a case-by-case basis to achieve greater performance than more rigid fixed policy approaches, using more computation when it is required by a more difficult task.

\section{Model}

We consider a class of fully observed, one-shot decision-making tasks (i.e., continuous, contextual bandits).
The performance objective is to find a \emph{control} $c \in \mathcal{C}$ which, given an initial state $x \in \mathcal{X}$, minimizes some loss function $\mathcal{L}$ between a known future goal state $x^*$ and the result of a forward process, $f(x, c)$.
The \emph{performance loss} $L_P$ is the (negative) utility of executing the control in the world, and is related to the optimal solution $c^*\in\mathcal{C}$ as follows:
\begin{align}
L_P(x^*, x, c) &= \mathcal{L}(x^*, f(x, c)), \label{eq:performance-loss}\\
c^* &= \argmin_c L_P(x^*, x, c). \label{eq:optimal-performance}
\end{align}
However, (\ref{eq:optimal-performance}) defines only the optimal solution---not how to achieve it.

\subsection{Optimizing Performance}

We consider an iterative optimization procedure that takes $x^*$ and $x$ as input and returns an approximation of $c^*$ in order to minimize (\ref{eq:performance-loss}).
The optimization procedure consists of a controller, which iteratively proposes controls, and an expert, which evaluates how good those controls are.
On the $n^\mathrm{th}$ iteration, the controller $\controller:\mathcal{X}\times \mathcal{X}\times \mathcal{H}\rightarrow \mathcal{C}$ takes as input, $x^*$, $x$, and information about the history of previously proposed controls and evaluations $h_{n-1}\in\mathcal{H}$, and returns a proposed control $c_n$ that aims to improve on previously proposed controls.
An expert $E:\mathcal{X}\times \mathcal{X}\times \mathcal{C}\rightarrow \mathcal{E}$ takes the proposed control and provides some information $e_n\in\mathcal{E}$ about the quality of the control, which we call an \emph{opinion}.
This opinion is added to the history, which is passed back to the controller, and the loop continues for $N$ steps, after which a final control $c_N$ is proposed.

Standard optimization methods use principled heuristics for proposing controls.
In gradient descent, for example, controls are proposed by adjusting $c_n$ in the direction of the gradient of the reward with respect to the control.
In Bayesian optimization, controls are proposed based on selection criteria such as ``probability of improvement'', or a meta-selection criterion for choosing among several basic selection criteria \cite{Hoffman2011,Shahriari2014}.
Rather than choosing one of several controllers, our work learns a single controller and instead focuses on selecting from multiple experts (see Sec. \ref{sec:opt-comp-cost}).
In some cases $f$ is known and inexpensive to compute, and thus the optimization procedure sets $E\equiv f$.
However, in many real-world settings, $f$ is expensive or non-stationary and so it can be advantageous to use an approximation of $f$ (e.g., a state transition model), $L_P$ (e.g., an action-value function), or any other quantity that gives some information about $f$ or $L_P$.

\subsection{Optimizing Computational Cost} \label{sec:opt-comp-cost}

\begin{figure*}[t!]
  \begin{center}
    \includegraphics[width=0.9\textwidth]{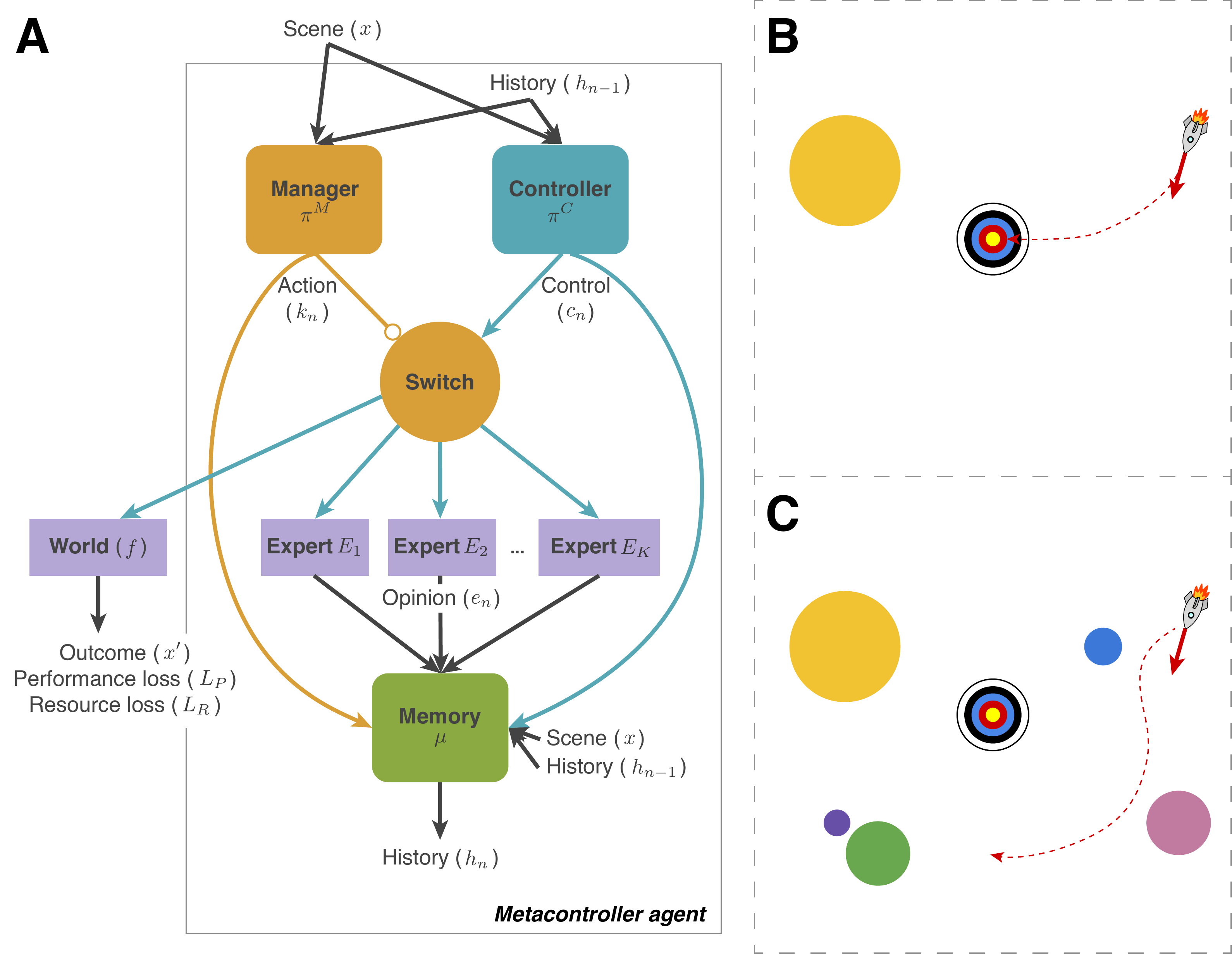}
    \caption{
    \textbf{Metacontroller architecture and task.}
    \textbf{A}: All components are part of the metacontroller agent (box) except the scene and the world, which are part of the agent's environment. 
    The \emph{manager} takes the scene and history and determines which action to take (i.e., whether to execute or ponder, and with what expert to ponder with), denoted by the orange lines.
    The \emph{controller} takes the scene and history and computes a control (e.g., the force to apply to a spaceship), denoted by the blue lines.
    The orange line ending with a circle at the switch reflects the fact that the manager's action affects the behavior of the switch, which routes the controller's control to either an \emph{expert} (e.g., a simulation model of the spaceship's trajectory, an action-value function, etc.) or the \emph{world}.
    The outcome and reward from the expert, along with the history, action, and control, are fed into the \emph{memory}, which produces the next history.
    The history is fed back to the controller on the next iteration in order to allow it to propose controls based on what it has already tried.
    \textbf{B-C}: Scenes consisted of a number of planets (depicted here by colored circles) of different masses as well as a spaceship (also with a variable mass). The task was to apply a force to the spaceship for one time step of simulation (depicted here as a solid red arrow) such that the resulting trajectory (dotted red arrow) would put the spaceship at a target (bullseye) after 11 steps of simulation. The white ring of the bullseye corresponds to a performance loss of 0.12-0.15, the black ring to a loss of 0.09-0.12, the blue ring to a loss of 0.06-0.09, the red ring to a loss of 0.03-0.06, and the yellow center to a loss of 0.03 or less. B depicts an easy, 1-planet scene, while C depicts a very difficult 5-planet scene.}
    \label{fig:model}
  \end{center}
\end{figure*}

Given a controller and one or more experts, there are two important decisions to be made.
First, how many optimization iterations should be performed?
The approximate solution usually improves with more iterations, but each iteration costs computational resources.
However, most traditional optimizers either ignore the cost of computation or select the number of iterations using simple heuristics.
Because they do not balance the cost of computation against the performance loss, the overall effectiveness of these approaches is subject to the skill and preferences of the practitioners who use them.
Second, which expert should be used on each step of the optimization?
Some experts may be accurate but expensive to compute in terms of time, energy and/or money, while others may be crude, yet cheap.
Moreover, the reliability of the experts may not be known \emph{a priori}, further limiting the effectiveness of the optimization procedure.
Our use of a metacontroller address these issues by jointly optimizing over the choices of how many steps to take and which experts to use.

We consider a family of optimizers which use the same controller, $\controller$, but vary in their expert evaluators, $\{E_1,\ldots{},E_K\}$.
Assuming that the controller and experts are deterministic functions, the number of iterations $N$ and the sequences of experts $\mathbf{k}=(k_1,\ldots{},k_{N-1})$ exactly determine the final control and performance loss $L_P$.
This means we have transformed the performance optimization over $c$ into an optimization over $N$ and $\mathbf{k}$:
$(N, \mathbf{k})^* = \argmin_{k, n} L_P(x^*, x, c(N, \mathbf{k}, x, x^*))$,
where the notation $c(N, \mathbf{k}, x, x^*)$ is used to emphasize that the control is a function $N$, $\mathbf{k}$, $x$, and $x^*$.

If each optimizer has an associated computational cost $\tau_k$,
then $N$ and $\mathbf{k}$ also exactly determine the computational \emph{resource loss} of the optimization run, $L_R(N,\mathbf{k})=\sum_{n=1}^{N-1}\tau_{k_n}$.
The total loss is then the sum of $L_P$ and $L_R$, each of which are functions of $N$ and $\mathbf{k}$,
\begin{align}
L_T(x^*, x, N, \mathbf{k}) &= L_P(x^*, x, c(N, \mathbf{k}, x, x^*)) + L_R(N, \mathbf{k}) \label{eq:total-loss} \\
&= \mathcal{L}(x^*, f(x, \controller(x^*, x, h_{N-1}))) + \sum_{n=1}^{N-1} \tau_{k_n},
\end{align}
and the optimal solution is defined as $(N, \mathbf{k})^* = \argmin_{N, \mathbf{k}} L_T(x^*, x, N, \mathbf{k})$.
Optimizing $L_T$ is difficult because of the recursive dependency on the history, $h_{N-1}$, and because the discrete choices of $N$ and $\mathbf{k}$ mean $L_T$ is not differentiable.

To optimize $L_T$ we recast it as an RL problem where the objective is to jointly optimize task performance and computational cost.
As shown in Figure~\ref{fig:model}a, the \textbf{metacontroller agent} $a^M$ is comprised of a controller $\controller$, a pool of experts $\{E_1, \dots, E_K\}$, a manager $\manager$, and a memory $\mu$.
The \emph{manager} is a meta-level policy \citep{Russel1991,Hay2012} over actions indexed by $k$, which determine whether to terminate the optimization procedure ($k=0$) or to perform another iteration of the optimization procedure with the $k^\textrm{th}$ expert.
Specifically, on the $n^\textrm{th}$ iteration the controller produces a new control $c_n$ based on the history of controls, experts, and evaluations.
The manager, also relying on this history, independently decides whether to end the optimization procedure (i.e., to execute the control in the world) or to perform another iteration and evaluate the proposed control with the $k_n^\textrm{th}$ expert (i.e., to \emph{ponder}, after \cite{Graves2016}).
The memory then updates the history $h_n$ by concatenating $k$, $c_n$, and $e_n$ with the previous history $h_{n-1}$.
Coming back to the notion of imagination-based optimization, we suggest that this iterative optimization process is analogous to imagining what will happen (using one or more approximate world models) before actually executing that action in the world.
For further details, see Appendix~\ref{sec:agents}, and for an algorithmic illustration of the metacontroller agent, see Algorithm~\ref{alg:metacontroller} in the appendix.

We also define two special cases of the metacontroller for baseline comparisons. 
The \textbf{iterative agent} $a^I$ does not have a manager and uses only a single expert. Its number of iterations are pre-set to a single $N$.
The \textbf{reactive agent}, $a^0$, is a special case of the iterative agent, where the number of iterations is fixed to $N=0$.
This implies that proposed controls are executed immediately in the world, and are not evaluated by an expert.
For algorithmic illustrations of the iterative and reactive agents, see Algorithms~\ref{alg:iterative} and \ref{alg:reactive} in the appendix.

\subsection{Neural Network Implementation}

We use standard deep learning building blocks, e.g., multi-layer perceptrons (MLPs), RNNs, etc., to implement the controller, experts, manager, and memory, because they are effective at approximating complex functions via gradient-based and reinforcement learning, but other approaches could be used as well.
In particular, we constructed our implementation to be able to make control decisions in complex dynamical systems, such as controlling the movement of a spaceship (Figure~\ref{fig:model}b-c), though we note that our approach is not limited to such physical reasoning tasks.
Here we used mean-squared error (MSE) for our $\mathcal{L}$ and Adam \citep{Kingma2014} as the training optimizer.

\paragraph{Experts}

We implemented the experts as MLPs and ``interaction networks'' (INs) \citep{Battaglia2016}, which are well-suited to predicting complex dynamical systems like those in our experiments below.
Each expert has parameters $\theta^{E_k}$, i.e. $e_n=E_k(x^*, x, c_n; \theta^{E_k})$, and may be trained either on-policy using the outputs of the controller (as is the case in this paper), or off-policy by any data that pairs states and controls with future states or reward outcomes.
The objective $L_{E_k}$ for each expert may be different depending on what the expert outputs.
For example, the objective could be the loss between the goal and future states, $L_{E_k} = \mathcal{L}\left(f(x, c), E_k(x^*, x, c; \theta^{E_k}) \right)$, which is what we use in our experiments.
Or, it could be the loss between $L_P$ and an action-value function that predicts $L_P$ directly, $L_{E_k} = \mathcal{L}\left( L_P(x^*, x, c), E_k(x^*, x, c; \theta^{E_k}) \right)$.
See Appendix~\ref{sec:experts} for details.

\paragraph{Controller and Memory}

We implemented the controller as an MLP with parameters $\theta^C$, i.e. $c_n=\controller(x^*, x, h_{n-1}; \theta^C)$, and we implemented the memory as a Long Short-Term Memory (LSTM) \citep{Hochreiter1997} with parameters $\theta^\mu$.
The memory embeds the history as a fixed-length vector, i.e. $h_n = \mu(h_{n-1}, k_n, c_n, E_{k_n}(x^*, x, c_n); \theta^\mu)$.
The controller and memory were trained jointly to optimize (\ref{eq:performance-loss}).
However, this objective includes $f$, which is often unknown or not differentiable.
We overcame this by approximating $L_P$ with a differentiable critic analogous to those used in policy gradient methods \citep[e.g.][]{Silver2014,Lillicrap2015,Heess2015}.
See Appendices~\ref{sec:critic} and~\ref{sec:controller-and-memory} for details.

\paragraph{Manager}

We implemented the manager as a stochastic policy that samples from a categorical distribution whose weights are produced by an MLP with parameters $\theta^M$, i.e. $k_n \sim \mathrm{Categorical}(k; \manager(x^*, x, h_{n-1}; \theta^M))$.
We trained the manager to minimize (\ref{eq:total-loss}) using \textsc{Reinforce} \citep{Williams1992}, but other deep RL algorithms could be used instead.
See Appendix~\ref{sec:manager} for details.

\section{Experiments}

To evaluate our metacontroller agent, we measured its ability to learn to solve a class of physics-based tasks that are surprisingly challenging.
Each episode consisted of a scene which contained a spaceship and multiple planets (Figure~\ref{fig:model}b-c).
The spaceship's goal was to rendezvous with its mothership near the center of the system in exactly 11 time steps, but it only had enough fuel to fire its thrusters once.
The planets were static but the gravitational force they exerted on the spacecraft induced complex non-linear dynamics on the motion over the 11 steps.
The spacecraft's action space was continuous, up to some maximum magnitude, and represented the instantaneous Cartesian velocity vector imparted by its thrusters.
Further details are in Appendix~\ref{sec:task}.

We trained the reactive, iterative, and metacontroller agents on five versions of the spaceship task involving different numbers of planets.\footnote{Available from: \texttt{https://www.github.com/deepmind/spaceship\_dataset}}
The iterative agent was trained to take anywhere from zero (i.e., the reactive agent) to ten ponder steps.
The metacontroller was allowed to take a maximum of ten ponder steps.
We considered three different experts which were all differentiable: an \emph{MLP expert} which used an MLP to predict the final location of the spaceship, an \emph{IN expert} which used an interaction network \citep{Battaglia2016} to predict the full trajectory of the spaceship, and a \emph{true simulation expert} which was the same as the world model.
In some conditions the metacontroller could use exactly one expert and in others it was allowed to select between the MLP and IN experts. 
For experiments with the true simulation expert, we used it to backpropagate gradients to the controller and memory. 
For experiments with an MLP as the only expert, we used a learned IN as the critic. For experiments with an IN as one of its experts, the critic was an IN with shared parameters.
We trained the metacontroller on a range of different ponder costs, $\tau_k$, for the different experts. 
Further details of the training procedure are available in Appendix~\ref{sec:implementation}.

\subsection{Reactive and iterative agents}

\begin{figure*}[t!]
  \begin{center}
    \includegraphics[width=\textwidth]{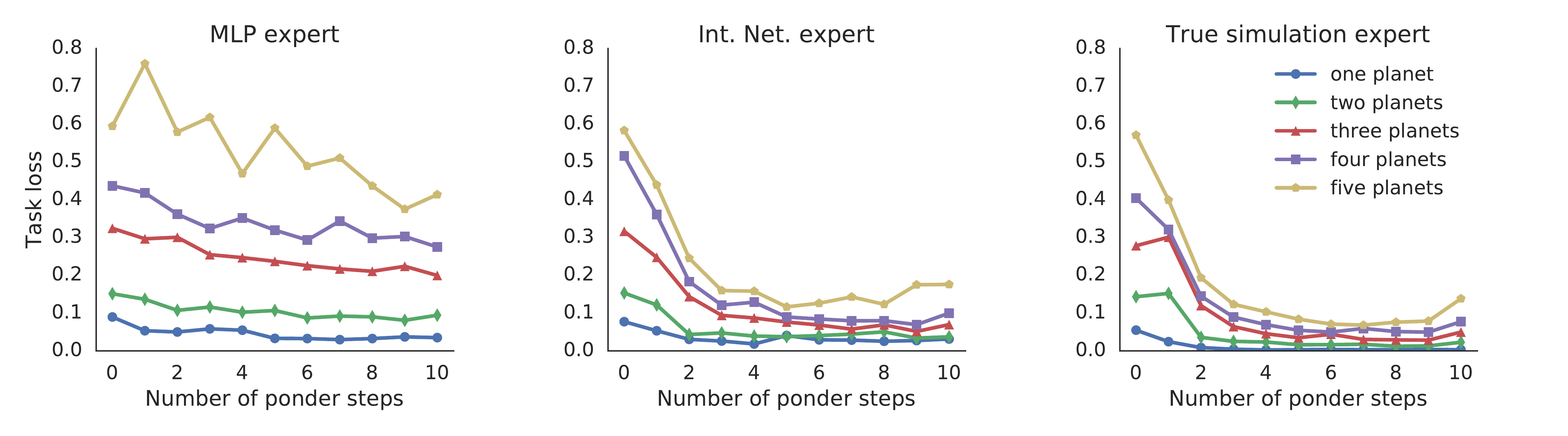}
    \caption{\textbf{Test performance of the reactive and iterative agents.} Each line corresponds to the performance of an iterative agent (either the true simulation expert, the MLP expert, or the interaction net expert) trained for a fixed number of ponder steps on one of the five datasets; the line color indicates which dataset the controller was trained on. In all cases, performance refers to the performance loss, $L_P$. \textbf{Left}: the MLP expert struggles with the task due to its limited expressivity, but still benefits from pondering. \textbf{Middle}: the IN expert performs almost as well as the true simulation expert, even though it is not a perfect model. \textbf{Right}: The true simulation expert does quite well on the task, especially with multiple ponder steps.}
    \label{fig:fixedlength}
  \end{center}
\end{figure*}

Figure~\ref{fig:fixedlength} shows the performance on the test set of the reactive and iterative agents for different numbers of ponder steps.
The reactive agent performed poorly on the task, especially when the task was more difficult. With the five planets dataset, it was only able to achieve a performance loss of $0.583$ on average (see Figure~\ref{fig:model} for a depiction of the magnitude of the loss). 
In contrast, the iterative agent with the true simulation expert performed much better, reaching ceiling performance on the datasets with one and two planets, and achieving a performance loss of $0.0683$ on the five planets dataset.
The IN and MLP experts also improve over the reactive agent, with a minimum performance loss of $0.117$ and $0.375$ on the five planets dataset, respectively.

Figure~\ref{fig:fixedlength} also highlights how important the choice of expert is.
When using the true simulation and IN experts, the iterative agent performs well.
With the MLP expert, however, performance is substantially diminished.
But despite the poor performance of the MLP expert, there is still some benefit of pondering with it. With even just a few steps, the MLP iterative agent outperforms its reactive counterpart.
However comparing the reactive agent with the $N=1$ iterative agent is somewhat unfair because the iterative agent has more parameters due to the expert and the memory.
However, given that there tends to \emph{also} be an increase in performance between one and two ponder steps (and beyond), it is clear that pondering---even with a highly inaccurate model---can still lead to better performance than a model-free reactive approach.

\subsection{Metacontroller with One Expert}

\begin{figure*}[t!]
  \begin{center}
    \includegraphics[width=\textwidth]{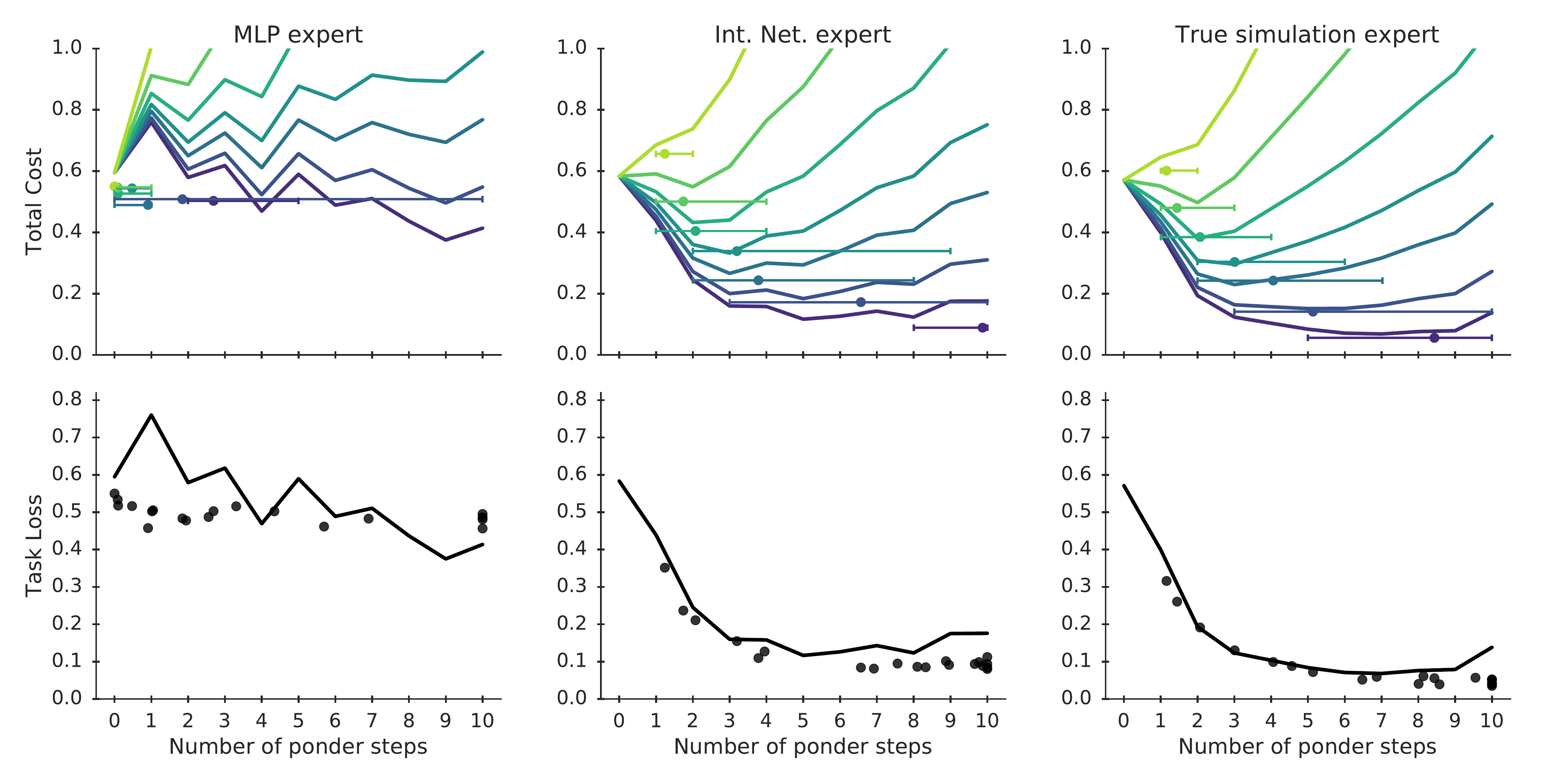}
    \caption{\textbf{Test performance of the metacontroller with a single expert on the five planets dataset.}
    Each column corresponds to a different experts.
    The lines indicate the performance of the iterative agents for different numbers of ponder steps.
    The points indicate the performance of the metacontroller, with each point corresponding to a different value of $\tau$.
    The $x$-coordinate of each point is an average across the number of ponder steps, and the $y$-coordinate is the average loss.
    \textbf{Top row:} Here we show total cost rather than just performance on the task (i.e., including computation cost). Different colors show the result for different $\tau$, with the different lines showing the cost for the same iterative controller under different values of $\tau$.
    The error bars (for the metacontroller) indicate 2.5\% and 97.5\% confidence intervals.
    When the point is below its corresponding curve, it means that the metacontroller was able to achieve a better speed-accuracy trade-off than that achievable by the iterative agent. 
    Line colors of increasing brightness correspond to increasing $\tau$, with $\tau$ values taken from $[0, 0.0134, 0.0354, 0.0576, 0.0934, 0.152, 0.246]$.
   \textbf{Bottom row:} Here we show just the performance loss (i.e., without computational cost). Each point corresponds to a different value of $\tau$. The fact that the points are below the curve means the metacontroller agent learns to perform better than the iterative agent with the equivalent number of ponder steps.}
    \label{fig:reinforce}
  \end{center}
\end{figure*}

Though the iterative agents achieve impressive results, they expend more computation than necessary.
For example, in the one and two planet conditions, the performances of the IN and true simulation iterative agents received little performance benefit from pondering more than two or three steps, while for the four and five planet conditions they required at least five to eight steps before their performance converged.
When computational resources have no cost, the number of steps are of no concern, but when they have some cost it is important to be economical.

Because the metacontroller learns to choose its number of pondering steps, it can balance its performance loss against the cost of computation.
Figure~\ref{fig:reinforce} (top row, middle and right subplots) shows that the IN and true simulation expert metacontroller take fewer ponder steps as $\tau$ increases, tracking closely the minimum of the iterative agent's cost curve (i.e., the metacontroller points are always near the iterative agent curves' minima). This adaptive behavior emerges automatically from the manager's learned policy, and avoids the need to perform a hyperparameter search to find the best number of iterations for a given $\tau$. 

\begin{figure*}[b!]
  \begin{center}
    \includegraphics[width=\textwidth]{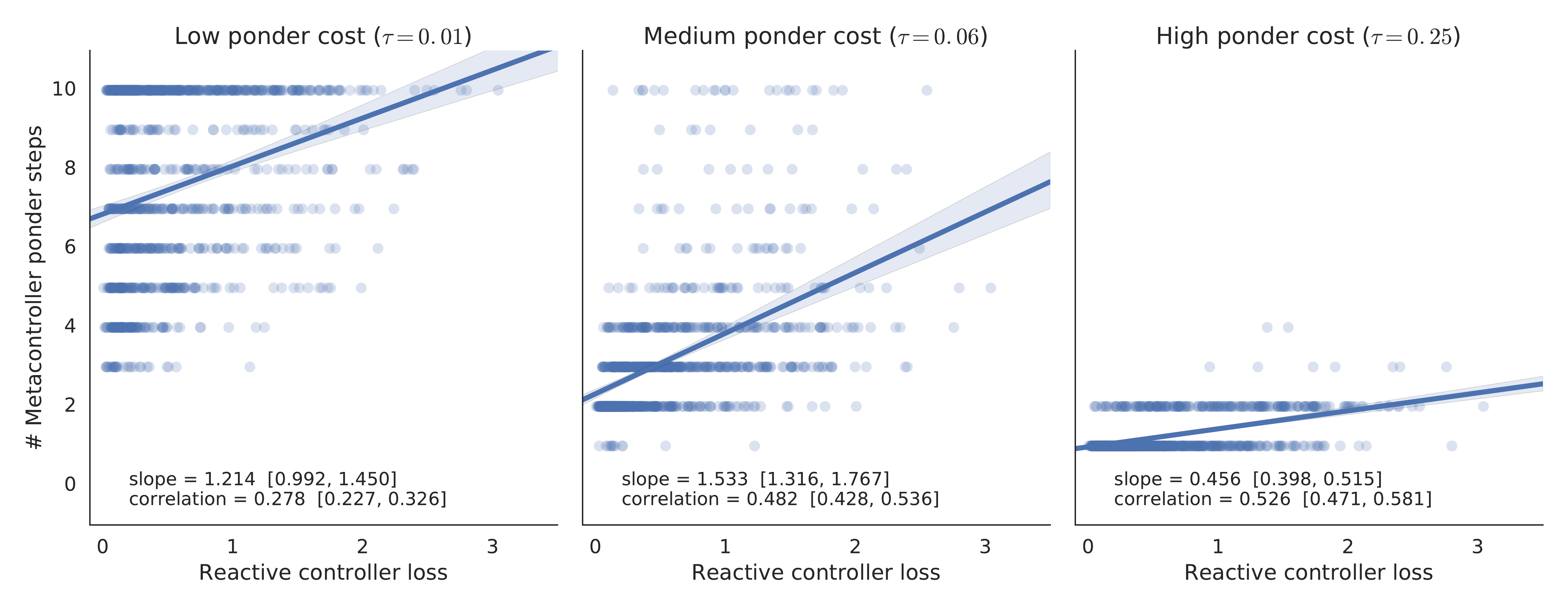}
    \caption{\textbf{Relationship between the number of ponder steps and per-episode difficulty for the IN metacontroller.} Each subplot's $x$-axis represents the episode difficulty, as measured by the reactive controller's loss. Each $y$-axis represents the number of ponder steps the metacontroller took. The points are individual episodes, and the line is the best fit regression line and 95\% confidence intervals. The different subplots show different values of $\tau$ (labeled in the title). In each case, there is a clear positive relationship between the difficulty of the task and the number of ponder steps, suggesting that the metacontroller learns to spend more time on hard problems and less time on easier problems. At the bottom of each plot are the fitted slope and correlation coefficient values, along with their 95\% confidence intervals in brackets.}
    \label{fig:reinforce-ponder-vs-reactive-loss}
  \end{center}
\end{figure*}

The metacontroller does not simply choose an average number of ponder steps to take per episode: it actually tailors this choice to the difficulty of each episode.
Figure~\ref{fig:reinforce-ponder-vs-reactive-loss} shows how the number of ponder steps the IN metacontroller chooses in each episode depends on that episode's difficulty, as measured by the episode's loss under the reactive agent.
For more difficult episodes, the metacontroller tends to take more ponder steps, as indicated by the positive slopes of the best fit lines, and this proportionality persists across the different levels of $\tau$ in each subplot.

The ability to adapt its choice of number of ponder steps on a per-episode basis is very valuable because it allows the metacontroller to spend additional computation only on those episodes which require it. The total costs of the IN and true simulation metacontrollers' are 11\% and 15\% lower (median) than the \emph{best achievable} costs of their corresponding iterative agents, respectively, across the range of $\tau$ values we tested (see Figure~\ref{fig:cost-scatter} in the Appendix for details).

There can even be a benefit to using a metacontroller when there are no computational resource costs. Consider the rightmost points in Figure~\ref{fig:reinforce} (bottom row, middle and right subplots), which show the performance loss for the IN and true simulation metacontrollers when $\tau$ is low.
Remarkably, these points still outperform the best achievable iterative agents.
This suggests that there can be an advantage to stopping pondering once a good solution is found, and more generally demonstrates that the metacontroller's learning process can lead to strategies that are superior to those available to less flexible agents.

The metacontroller with the MLP expert had very poor average performance and high variance on the five planet condition (Figure~\ref{fig:reinforce}, top left subplot), which is why we restricted our focus in this section to how the metacontrollers with IN and true simulation experts behaved. The MLP's poor performance is crucial, however, for the following section (\ref{sec:results-metacontroller-2}) which analyzes how a multiple-expert metacontroller manages experts which vary greater in their reliability.

\subsection{Metacontroller with Two Experts} \label{sec:results-metacontroller-2}

\begin{figure*}[t!]
  \begin{center}
    \includegraphics[width=0.6\textwidth]{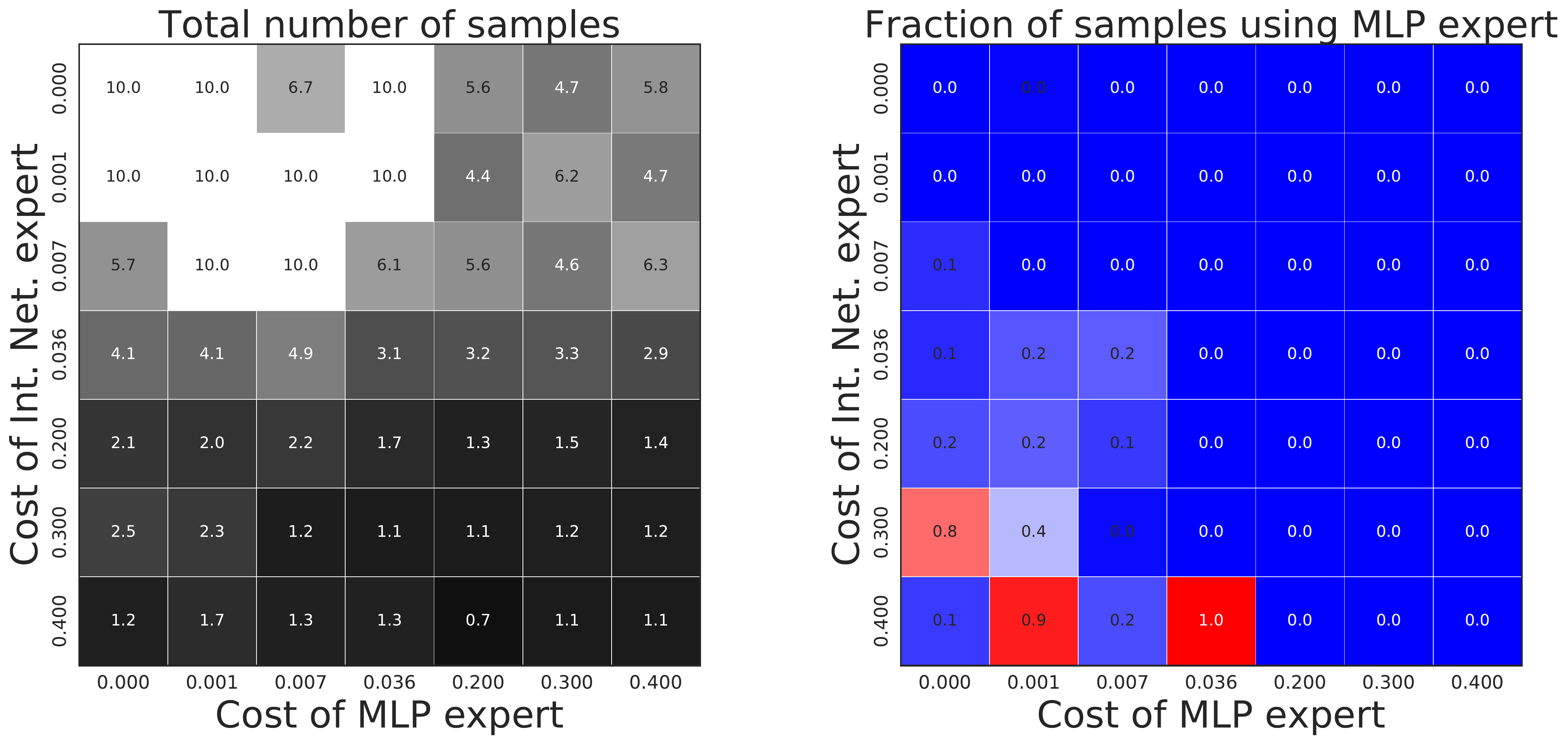}%
    \caption{\textbf{Test performance of the metacontroller with multiple experts on the five planets dataset.} \textbf{Left}: The average number of total ponder steps, for different values of $\tau$.  As with the single-expert metacontrollers, fewer ponder steps are taken when the cost is very high, and more are taken when the cost is low. \textbf{Right}: The fraction of ponder steps taken by the MLP expert relative to the IN expert. In the majority of cases, the metacontroller favors using the IN expert as it is much more reliable. The few exceptions (red squares) are cases when the cost of the IN expert is much higher relative to the cost of the MLP expert.
}
    \label{fig:reinforce3}
  \end{center}
\end{figure*}

When we allow the manager to additionally choose between two experts, rather than only relying on a single expert, we find a similar pattern of results in terms of the number of ponder steps (Figure~\ref{fig:reinforce3}, left).
Additionally, the metacontroller is successfully able to identify the more reliable IN network and consequently uses it a majority of the time, except in a few cases where the cost of the IN network is extremely high relative to the cost of the MLP network (Figure~\ref{fig:reinforce3}, right).
This pattern of results makes sense given the good performance (described in the previous section) of the metacontroller with the IN expert compared to the poor performance of the metacontroller with the MLP expert.
The manager \emph{should not} generally rely on the MLP expert because it is simply not a reliable source of information.

However, the metacontroller has more difficulty finding an optimal balance between the two experts on a step-by-step basis: the addition of a second expert did not yield much of an improvement over the single-expert metacontroller, with only $9$\% of the different versions (trained with different $\tau$ values for the two experts) achieving a lower loss than the best iterative controller.
We believe the mixed performance of the metacontroller with multiple experts is partially due to an entropy term which we used to encourage the manager's policy to be non-deterministic (see Appendix~\ref{sec:manager}).
In particular, for high values of $\tau$, the optimal thing to do is to always execute immediately without pondering.
However, because of the entropy term, the manager is encourage to have a non-deterministic policy and therefore is likely to ponder more than it should---and to use experts that are more unreliable---even when this is suboptimal in terms of the total loss (\ref{eq:total-loss}).

Despite the fact that the metacontroller with multiple experts does not result in a substantial improvement over that which uses a single expert, we emphasize that the manager is able to identify and use the more reliable expert the majority of the time.
And, it is still able to choose a variable number of steps according to how difficult the task is (Figure~\ref{fig:reinforce3}, left).
This, in and of itself, is an improvement over more traditional optimization methods which would require that the expert is hand-picked ahead of time and that the number of steps are determined heuristically.

\section{Discussion}

In this paper, we have presented an approach to adaptive, imagination-based optimization in neural networks.
Our approach is able to flexibly choose \emph{which} computations to perform as well as \emph{how many} computations need to be performed, approximately solving a speed-accuracy trade-off that depends on the difficulty of the task.
In this way, our approach learns to rely on whatever source of information is most useful \emph{and} most efficient.
Additionally, by consulting the experts on-the-fly, our approach allows agents to test out actions to ensure that their consequences are not disastrous before actually executing them.

While the experiments in this paper involve a one-shot decision task, our approach lays a foundation that can be built upon to support more complex situations.
For example, rather than applying a force only on the first time step, we could turn the problem into one of trajectory optimization for continuous control by asking the controller to produce a sequence of forces.
In the case of planning, our approach could potentially be combined with methods like Monte Carlo Tree-Search (MCTS) \citep{Coulom2006}, where our experts would be akin to having several different rollout policies to choose from, and our controller would be akin to the tree policy.
While most MCTS implementations will run rollouts until a fixed amount of time has passed, our approach would allow the manager to adaptively choose the number of rollouts to perform and which policies to perform the rollouts with.
Our method could also be used to naturally augment existing model-free approaches such as DQN \citep{Mnih2015} with online model-based optimization by using the model-free policy as a controller and adding additional experts in the form of state-transition models.
An interesting extension would be to compare our metacontroller architecture with a na\"ive model-based controller that performs gradient-based optimization to produce the final control. We expect our metacontroller architecture might require fewer model evaluations and to be more robust to model inaccuracies compared to the gradient-based method, because our method has access to the full history of proposed controls and evaluations whereas traditional gradient-based methods do not.

Although we rely on differentiable experts in our metacontroller architecture, we do not utilize the gradient information from these experts.
An interesting extension to our work would be to pass this gradient information through to the manager and controller (as in \cite{Andrychowicz2016}), which would likely improve performance further, especially in the more complex situations discussed here.
Another possibility is to train some or all of the experts inline with the controller and metacontroller, rather than independently, which could allow their learned functionality to be more tightly integrated with the rest of the optimization loop, at the expense of their generality and ability to be repurposed for other uses.

To conclude, we have demonstrated how neural network-based agents can use metareasoning to adaptively choose what to think about, how to think about it, and for how long to think for.
Our method is directly inspired by human cognition and suggests a way to make agents much more flexible and adaptive than they currently are, both in decision making tasks such as the one described here, as well as in planning and control settings more broadly.

\section*{Acknowledgments}

We would like to thank Matt Hoffman, Andrea Tacchetti, Tom Erez, Nando de Freitas, Guillaume Desjardins, Joseph Modayil, Hubert Soyer, Alex Graves, David Reichert, Theo Weber, Jon Scholz, Will Dabney, and others on the DeepMind team for helpful discussions and feedback.

{\footnotesize
\setlength{\bibsep}{2pt}
\bibliography{references}
\bibliographystyle{iclr2017_conference}}

%%%%%%%%%%%%%%%%%%%%%%%%%%%%%%%%%%%%%%%%%%%%%%%%%%%%%%%%%
%%%%%%%%%%%%%%%%%%%%%%%%%%%%%%%%%%%%%%%%%%%%%%%%%%%%%%%%%
%%%%%%%%%%%%%%%%%%%%%%%%%%%%%%%%%%%%%%%%%%%%%%%%%%%%%%%%%

\clearpage{}
\appendix

\section{Metacontroller Details}
\label{sec:agents}

Here, we give the precise definitions of the metacontroller agent.
As described in the main text, the iterative and reactive agents are special cases of the metacontroller agent, and are therefore not discussed here.

The metacontroller agent $a^M$ is comprised of the following components:

\begin{itemize}
\item A \emph{history-sensitive controller}, $\controller: \mathcal{X} \times \mathcal{X} \times \mathcal{H} \rightarrow \mathcal{C}$, which is a policy that maps goal and initial states, and a history, $h \in \mathcal{H}$, to controls, whose aim is to minimize (\ref{eq:performance-loss}).

\item A \emph{pool of experts} $\{E_1, \dots, E_K\}$. Each expert $E: \mathcal{X} \times \mathcal{X} \times \mathcal{C} \rightarrow \mathcal{E}$ maps goal states, input states, and actions to \emph{opinions}.
Opinions can be either states-only ($\mathcal{E} = \mathcal{X}$), states and rewards ($\mathcal{E} = \mathcal{X} \times \mathbb{R}$), or rewards-only ($\mathcal{E} = \mathbb{R}$).
The expert corresponds to the evaluator for the optimization routine, i.e., an approximation of the forward process $f$.

\item A \emph{manager}, $\manager: \mathcal{X} \times \mathcal{X} \times \mathcal{H}_n \rightarrow \{0, \dots, K\}$, which is a policy which decides whether to send a proposed control to the world ($k=0$) or to the $k^\mathrm{th}$ expert for evaluation, in order to minimize (\ref{eq:total-loss}).
This formulation is based on that used by metareasoning systems \citepappendix{Russel1991,Hay2012}.
Details on the corresponding MDP are given in Appendix~\ref{sec:mdp}.

\item A \emph{memory}, $\mu: \mathcal{H}_{n - 1} \times \mathcal{Z} \rightarrow \mathcal{H}_n$, which is a function that maps the prior history $h_{n-1} \in \mathcal{H}_{n-1}$, as well as the most recent manager choice, proposed control, and expert evaluation $(k, c, e) \in \{0, \dots, K\} \times \mathcal{C} \times \mathcal{E} = \mathcal{Z}$, to an updated history $h_n \in \mathcal{H}_n$, which is then made available to the manager and controller on subsequent iterations. 
The history at step $n$ is a recursively defined tuple which is the concatenation of the prior history with the most recently proposed control, expert evaluation, and expert identity: $h_n = h_{n-1} \cap \left((k_n, c_n, E_{k_n}(x^*, x, c_n))\right) = \left((k_1, c_1, E_{k_1}(x^*, x, c_1)), \dots, (k_n, c_n, E_{k_n}(x^*, x, c_n))\right)$ where $h_0 = ()$ represents an empty initial history.
Similarly, the finite set of histories up to step $n$ is: $\mathcal{H}_n = \mathcal{H}_{n-1} \times \mathcal{Z} = \mathcal{Z}^n$ where $\mathcal{H}_0 = \{()\}$.
\end{itemize}

The metacontroller produces:
\begin{equation}
a^M(x^*, x) = \controller(x^*, x, h_{N-1}) = c_N \label{eq:metacontroller-agent}
\end{equation}
where $N = n \mbox{\; s.t.\; } k_n = 0$. This function is summarized in Algorithm~\ref{alg:metacontroller}. The other agents (iterative and reactive), as mentioned in the main text, are simpler versions of the metacontroller agent and are summarized in Algorithms~\ref{alg:iterative} and \ref{alg:reactive}.

\subsection{Meta-Level MDP}
\label{sec:mdp}

To implement the manager for the metacontroller agent, we draw inspiration from the metareasoning literature \citepappendix{Russel1991,Hay2012} and formulate the problem as a finite-horizon Markov Decision Process (MDP) $\langle \mathcal{S}, \mathcal{A}, P, R\rangle$ over the decision of whether to perform another iteration of the optimization procedure or to execute a control in the world.
\begin{itemize}
\item The state space $\mathcal{S}$ consists of goal states, external states, and internal histories, $\mathcal{S} = \mathcal{X} \times \mathcal{X} \times \mathcal{H}$.
\item The action space $\mathcal{A}$ contains $K+1$ discrete actions, $\{0, \dots, K\}$, which correspond to \emph{execute} ($k = 0$) and \emph{ponder} ($k \in \{1, \dots, K\}$), where \emph{ponder} (after \citetappendix{Graves2016}) refers to performing an iteration of the optimization procedure with the $k^\mathrm{th}$ expert.
\item The (deterministic) state transition model $P: \mathcal{S} \times \mathcal{C} \times \mathcal{S} \rightarrow [0, 1]$ is, 
\begin{equation*}
P(x^\prime, h_{n} | x^*, x, h_{n-1}, k) = 
\begin{cases}
P(x^\prime | x^*, x, h_{n-1}, k) & \mbox{ if } k = 0 \\
P(h_{n} | x^*, x, h_{n-1}, k) & \mbox{ otherwise}
\end{cases}
\end{equation*}
where $x^\prime = f(x, c)$ and $c = \controller(x^*, x, h_{n-1})$ and,
\begin{align*}
P(x^\prime | x^*, x, h_{n-1}, k) &= 
\begin{cases}
1 & \mbox{ if } x^\prime = f(x, c) \\
0 & \mbox{ otherwise}
\end{cases}
\\
P(h_{n} | x^*, x, h_{n-1}, k) &= 
\begin{cases}
1 & \mbox{ if } h_n = h_{n-1} \cup \{(k, c, E_{k}(x^*, x, c))\} \\
0 & \mbox{ otherwise}
\end{cases}
\end{align*} 
\item The (deterministic) reward function $R: \mathcal{S} \times \mathcal{A} \times \mathcal{S} \rightarrow \mathbb{R}$ maps the current state, current action, and next state to real-valued loss:
\begin{equation*}
R(x^*, x, h_{n-1}, k, x^\prime) = \begin{cases}
\mathcal{L}(x^*, x^\prime) & \mbox{ if } k = 0 \mbox{ (see Eq. \ref{eq:performance-loss})} \\
\tau_k & \mbox{ otherwise (see Eq. \ref{eq:total-loss})}
\end{cases}
\end{equation*}
where $x^\prime = f\left(x, \controller(x^*, x, h_{n-1})\right)$.
\end{itemize}

We approximate the solution to this MDP with a stochastic manager policy $\manager$. The manager chooses actions proportional to the immediate reward for taking action $k$ in state $s_n$ plus the expected sum of future rewards. This construction imposes a trade-off between accuracy and resources, incentivizing the agent to ponder longer and with more accurate (and potentially expensive) experts when the problem is harder.

\renewcommand{\Comment}[2][.5\linewidth]{%
  \leavevmode\hfill\makebox[#1][l]{$\triangleright$ #2}}

\begin{algorithm}[t]
\begin{algorithmic}[1]
\Function{$a^M$}{$x, x^*$}
  \State $h_0\gets ()$\Comment{Initial empty history}
  \State $k_0\gets \pi^M(x, x^*, h_0)$\Comment{Get an action from the manager}
  \State $c_0\gets \pi^C(x, x^*, h_0)$\Comment{Propose a control with the controller}
  \State $n\gets 0$
  \While {$k_n\neq 0$}\Comment{When $k\neq0$, ponder with an expert}
      \State $e_n\gets E_{k_n}(x, x^*, c_n)$\Comment{Get an expert's opinion}
      \State $h_{n+1}\gets \mu(h_n, k_n, c_n, e_n)$\Comment{Update the history}
      \State $n\gets n+1$
      \State $k_n\gets \pi^M(x, x^*, h_n)$\Comment{Choose the next action}
      \State $c_n\gets \pi^C(x, x^*, h_n)$\Comment{Propose the next control}
  \EndWhile
  \State\Return $c_n$
\EndFunction
\end{algorithmic}
\caption{Metacontroller agent. $x$ is the scene and $x^*$ is the target.}
\label{alg:metacontroller}
\end{algorithm}

\begin{algorithm}[t]
\begin{algorithmic}[1]
\Function{$a^I$}{$x, x^*, N$}
  \State $h_0\gets ()$\Comment{Initial empty history}
  \State $c_0\gets \pi^C(x, x^*, h_0)$\Comment{Propose a control with the controller}
  \State $n\gets 0$
  \While {$n< N$}\Comment{Ponder with an expert for $N$ steps}
      \State $e_n\gets E(x, x^*, c_n)$\Comment{Get the expert's opinion}
      \State $h_{n+1}\gets \mu(h_n, k_n, c_n, e_n)$\Comment{Update the history}
      \State $n\gets n+1$
      \State $c_n\gets \pi^C(x, x^*, h_n)$\Comment{Propose the next control}
  \EndWhile
  \State\Return $c_n$
\EndFunction
\end{algorithmic}
\caption{Iterative agent. $x$ is the scene, $x^*$ is the target, and $N$ is the number of ponder steps.}
\label{alg:iterative}
\end{algorithm}

\begin{algorithm}[t!]
\begin{algorithmic}[1]
\Function{$a^0$}{$x, x^*$}
  \State $c_0\gets \pi^C(x, x^*, ())$\Comment{Propose a control with the controller}
  \State\Return $c_0$
\EndFunction
\end{algorithmic}
\caption{Reactive agent. $x$ is the scene and $x^*$ is the target.}
\label{alg:reactive}
\end{algorithm}

\section{Gradients}
\label{sec:gradients}

\begin{figure}[t!]
\centering
\includegraphics[width=0.9\textwidth]{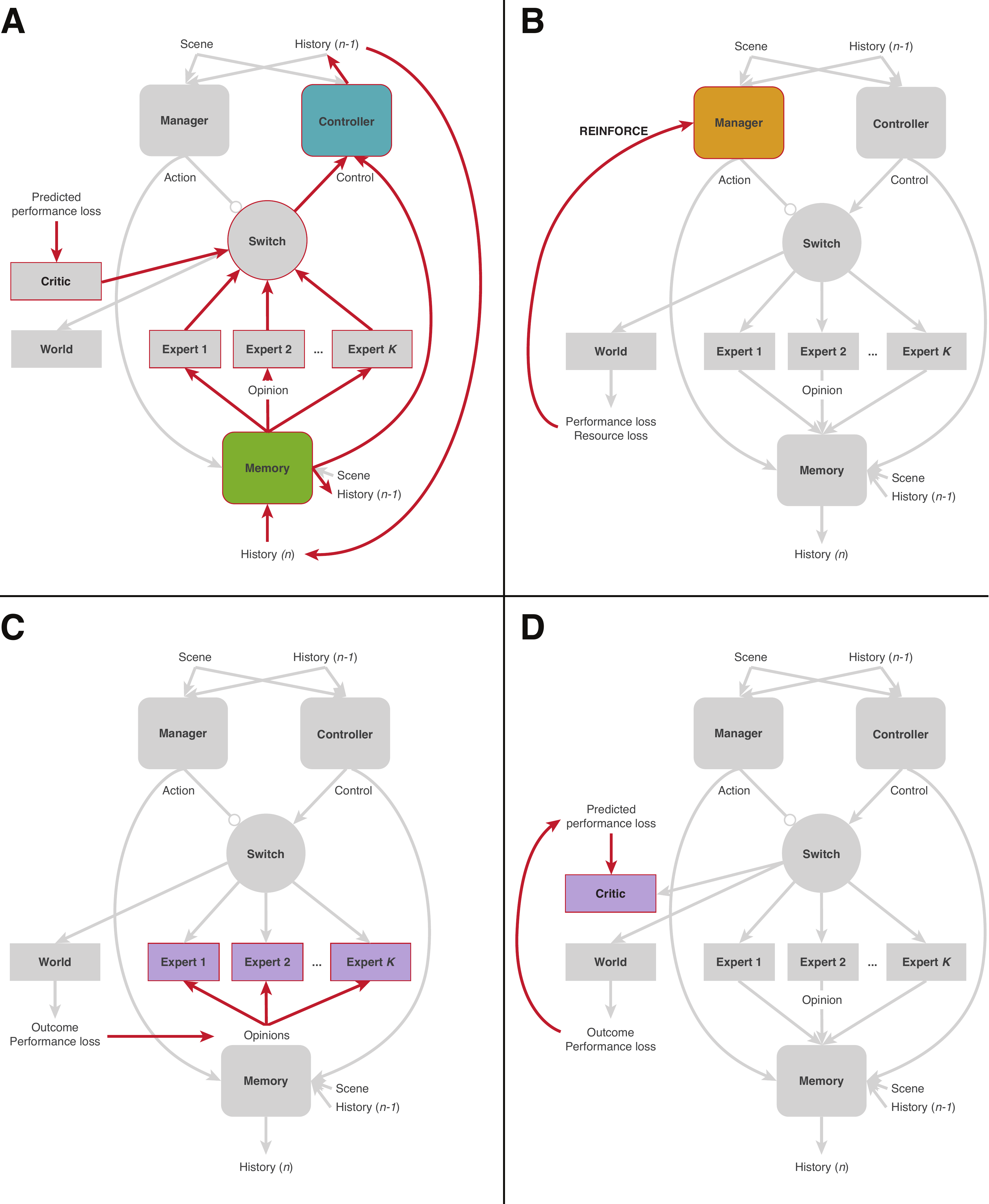}
\caption{\textbf{Training each part of the network.} In each subplot, red arrows depict gradients. Dotted arrows indicate backward connections that are not part of the forward pass. Colored nodes indicate weights that are being updated. All backpropagation occurs at the very end of a full forward pass (i.e., after the control has been executed in the world). \textbf{A}: Training the controller and memory with backpropagation-through-time (BPTT), beginning with the critic, and flowing to the controller, through the memory, through the relevant expert, through the controller again, and so on. \textbf{B}: Training the manager using \textsc{Reinforce} \citep{Williams1992}. \textbf{C}: Training the experts (note that each expert may have a different loss with respect to the outcome from the world). \textbf{D}: Training the critic.}
\label{fig:backprop}
\end{figure}

\subsection{Experts}
\label{sec:experts}

Training the experts is a straightforward supervised learning problem (Figure~\ref{fig:backprop}c).
The gradient is:
\begin{equation}
\frac{\partial \mathcal{L}^{E_k}}{\partial \theta^{E_k}}=\frac{\partial \mathcal{L}^{E_k}}{\partial E_k}\frac{\partial E_k}{\partial \theta^{E_k}},
\end{equation}
where $E_k$ is the $k^\mathrm{th}$ expert and $\mathcal{L}^{E_k}$ is the loss function for the $k^\mathrm{th}$ expert.
For example, in the case of an action-value function expert, this loss function might be $\mathcal{L}^{E_k}(f, E_k)=\left\Vert \mathcal{L}(x^*, f(x, c)) - E_k(x^*, x, c; \theta^{E_k})\right\Vert_2$.
In the case of an expert that predicts the final state using a model of the system dynamics, it might be $\mathcal{L}^{E_k}(f, E_k)=\left\Vert f(x, c) - E_k(x^*, x, c; \theta^{E_k})\right\Vert_2$.

\subsection{Critic}
\label{sec:critic}

The critic, $\hat{L}_P$, is an approximate model of the performance loss, $L_P$, (\ref{eq:performance-loss}), which is used to backpropagate gradients to the controller and memory. This means the critic can either be an action-value function, which approximates $\hat{L}_P = E_0 \approx L_P$ directly, or a model of the system dynamics composed with a known loss function between the goal and future states, $\hat{L}_P = \mathcal{L} \circ E_0 \approx \mathcal{L} \circ f$.
We train the critic, $E_0: \mathcal{X} \times \mathcal{X} \times \mathcal{C} \rightarrow \mathbb{R}$, using the same procedure as the experts are trained (Figure~\ref{fig:backprop}d).
A good expert may even be used as the critic.

\subsection{Controller and Memory}
\label{sec:controller-and-memory}

As shown in Figure~\ref{fig:backprop}a, we trained the controller and memory using backpropagation through time (BPTT) with an actor-critic architecture.
Specifically, rather than assuming $f$ is known and differentiable, we use a critic and backpropagate through it \citepappendix{Heess2015}:
\begin{align}
\frac{\partial \mathcal{L}}{\partial \theta^C}=\frac{\partial \mathcal{L}}{\partial E_*}\frac{\partial E_*}{\partial \controller_n}\frac{\partial \controller_n}{\partial \mu_n}\frac{\partial^+ \mu_n}{\partial \controller_{n-1}}\cdots{}\frac{\partial \controller_0}{\partial \theta^C}, & &
\frac{\partial \mathcal{L}}{\partial \theta^\mu}=\frac{\partial \mathcal{L}}{\partial E_*}\frac{\partial E_*}{\partial \controller_n}\frac{\partial \controller_n}{\partial \mu_n}\frac{\partial^+ \mu_n}{\partial \mu_{n-1}}\cdots{}\frac{\partial \mu_0}{\partial \theta^\mu}
\end{align}
where $E_*$ is the critic, $n$ is the maximum number of iterations the controller can use, and:
\begin{align}
\frac{\partial^+ \mu_n}{\partial \controller_{n-1}}=\frac{\partial \mu_n}{\partial E_{k_{n-1}}}\frac{\partial E_{k_{n-1}}}{\partial \controller_{n-1}}+\frac{\partial \mu_n}{\partial \controller_{n-1}}, & &
\frac{\partial^+ \mu_n}{\partial \mu_{n-1}}=\frac{\partial^+ \mu_n}{\partial \controller_{n-1}}+\frac{\partial \mu_n}{\partial \mu_{n-1}}
\end{align}
where we are using the $\partial^+$ notation to indicate summed gradients, following \citetappendix{Pascanu2013}.
Since $k_n$ has already been produced by the manager it can be treated as a constant and will produce an unbiased estimate of the gradient. This is convenient because it allows for training the controller and manager separately, or testing the controller's behavior with arbitrary actions post-training.

\subsection{Manager}
\label{sec:manager}

As discussed in the main text, we used the \textsc{Reinforce} algorithm \citetappendix{Williams1992} to train the manager (Figure~\ref{fig:backprop}b).
One potential issue, however, is that when training the controller and manager simultaneously, the controller will result in high cost early on in training and thus the manager will learn to always choose the \emph{execute} action. To discourage the manager from learning what is an essentially deterministic policy, we included a regularization term based on the entropy, $L_H$ \citepappendix{Williams1991,Mnih2016}:
\begin{align*}
L_H(\:\cdot{}\:;\theta^M) &= \lambda \; \mathbb{E}_{\manager}[\log \manager(\:\cdot{}\:;\theta^M)] \\
\frac{\partial \mathbb{E}_{\manager}[r]}{\partial \theta^M} &= \left(r - L_H(\:\cdot{}\:;\theta^M)\right) \; \frac{\partial}{\partial \theta^M}\log \manager(\:\cdot{}\:;\theta^M),
\end{align*}
$r$ is the full return given by (\ref{eq:total-loss}) and $\lambda$ is the strength of the regularization term.

\section{Spaceship Task}
\label{sec:task}

\subsection{Datasets}

We generated five datasets, each containing scenes with a different number of planets (ranging from a single planet to five planets).
Each dataset consisted of 100,000 training scenes and 1,000 testing scenes.
The target in each scene was always located at the origin, and each scene always had a sun with a mass of 100 units.
The sun was located between 100 and 200 distance units away from the target, with this distance sampled uniformly at random.
The other planets had a mass between 20 and 50 units, and were located 100 to 250 distance units away from the target, sampled uniformly at random.
The spaceship had a mass between 1 and 9 units, and was located 150 to 250 distance units away from the target.
The planets were always fixed (i.e., they could not move), and the spaceship always started at the beginning of each episode with zero velocity.

\subsection{Environment}

We simulated our scenes using a physical simulation of gravitational dynamics.
The planets were always stationary (i.e., they were not acted upon by any of the objects in the scene) but acted upon the spaceship with a force of:
\begin{equation}
\mathbf{F}_p = G\frac{m_pm_s}{r ^ 3}(\mathbf{x}_p - \mathbf{x}_s),
\end{equation}
where $\mathbf{F}_p$ is the force vector of the planet on the spaceship, $G=1000000$ is a gravitational constant, $m_p$ is the mass of the planet, $m_s$ is the mass of the spaceship, $r$ is the distance between the centers of masses of the planet and the spaceship, $\mathbf{x}_p$ is the location of the planet, and $\mathbf{x}_s$ is the location of the spaceship.
We simulated this environment using the Euler method, i.e.:
\begin{align}
\mathbf{a}_s = \frac{(\sum_{p}\mathbf{F}_p) - d\mathbf{v}_s + \mathbf{c}}{m_s} & &
\mathbf{x}_s^\prime = \mathbf{x}_s + \epsilon\mathbf{v}_s & &
\mathbf{v}_s^\prime = \mathbf{v}_s + \epsilon\mathbf{a}_s
\end{align}
where $\mathbf{a}_s$, $\mathbf{v}_s$, and $\mathbf{x}_s$ are the acceleration, velocity, and position of the spaceship, respectively; $d=0.1$ is a damping constant; $\mathbf{c}$ is the control force applied to the spaceship; and $\epsilon$ is the step size.
Note that we set $\mathbf{c}$ to zero for all timesteps except the first.

\section{Implementation Details}
\label{sec:implementation}

We used TensorFlow \citepappendix{TensorFlow} to implement and train all versions of the model.

\subsection{Architecture}

In our implementation of the controller, we used a two-layer MLP each with 100 units.
The first layer used ReLU activations and the second layer used a multiplicative interaction similar to \citetappendix{VandenOord2016}, which we found to work better in practice.
In our implementation of the memory, we used a single LSTM layer of size 100.
In our implementation of the manager, we used a MLP of two fully connected layers of 100 units each, with ReLU nonlinearities.

We constructed three different experts to test the various controllers.
The \emph{true simulation} expert was the same as the world model, and consisted of a simulation for 11 timesteps with $\epsilon=0.05$ (see Appendix~\ref{sec:task}).
The IN expert was an interaction network \citepappendix{Battaglia2016}, which has previously been shown to be able to learn to predict $n$-body dynamics accurately for simple systems.
The IN consists of a relational module and an object module.
In our case, the relational module was composed of 4 hidden layers of 150 nodes each, outputting ``effects'' encodings of size 100.
These effects, together with the relational model input are then used as input to the object model, which contained a single hidden layer of 100 nodes.
The object model outputs the velocity of the spaceship and we trained it to predict the velocity on every timestep of the spaceship's trajectory.
The \emph{MLP} expert was a MLP that predicted the final location of the spaceship and had the same architecture as the controller.

As discussed in Appendix~\ref{sec:gradients}, we used a critic to train the controller and memory.
We always used the IN expert as the critic, except in the case when the true simulation expert was used, in which case we also used the true simulation as the critic.

\subsection{Training Procedure}

All weights were initialized uniformly at random between 0 and 0.01.
An iteration of training consisted of gradient updates over a minibatch of size 1000; in total, we ran training for 100,000 iterations.
We additionally used a waterfall schedule for each of the learning rates during training, such that after 1000 iterations, if the loss was not decreasing, we would decay the step size by 5\%.

We trained the controller and memory together using the Adam optimizer \citepappendix{Kingma2014} with gradients clipped to a maximum global norm of 10 \citepappendix{Pascanu2013}.
The manager was trained simultaneously, but using a different learning rate than the controller and memory.
The IN and MLP experts were also trained simultaneously, but again with different learning rates.
Learning rates were determined using a grid search over a small number of values, and are given in Table~\ref{tbl:iterative} for the iterative agent, in Table~\ref{tbl:meta-one-expert} for the metacontroller with one expert, and in Table~\ref{tbl:meta-two-experts} for the metacontroller with two experts.

The iterative agent was trained to take a fixed number of ponder steps, ranging from 0 (i.e., the reactive agent) to 10.
The metacontrollers were allowed to take a variable number of ponder steps up to a maximum of 10.
For the metacontroller with a single expert, we trained the manager using $\tau=0$ and 20 additional values of $\tau$ spaced logarithmically between 0.00004 and 0.4 (inclusive).
For the metacontroller with multiple experts, we trained the manager on a grid of pairs of $\tau$ values, where each expert could have $\tau=0$ or one of 6 values spaced logarithmically between 0.00004 and 0.2 (inclusive).
In all cases, the entropy penalty for the metacontroller was $\lambda=0.2$.

\subsection{Convergence}

\paragraph{Reactive agent.} Training for the reactive agents was straightforward and converged reliably on all datasets.

\paragraph{Iterative agent.} For the iterative agent with the interaction network or true simulation experts, convergence was also reliable for small numbers of ponder steps.
Convergence was somewhat less reliable for larger numbers of ponder steps.
We believe this is because for some scenes, a larger number of ponder steps was more than necessary to solve the task (as is evidenced by the plateauing performance in Figure~\ref{fig:fixedlength}).
So, the iterative agent had to effectively ``remember'' what the best control was while it took the last few ponder steps, which is a more complicated and difficult task to perform.

For the iterative agent with the MLP expert, convergence was more variable especially when the task was harder, as can be seen in the variable performance on the five planets dataset in Figure~\ref{fig:fixedlength} (left).
We believe this is because the MLP agent was so poor, and that convergence would have been more reliable with a better agent.

\begin{figure}[t!]
\centering
\includegraphics[height=0.5\textwidth]{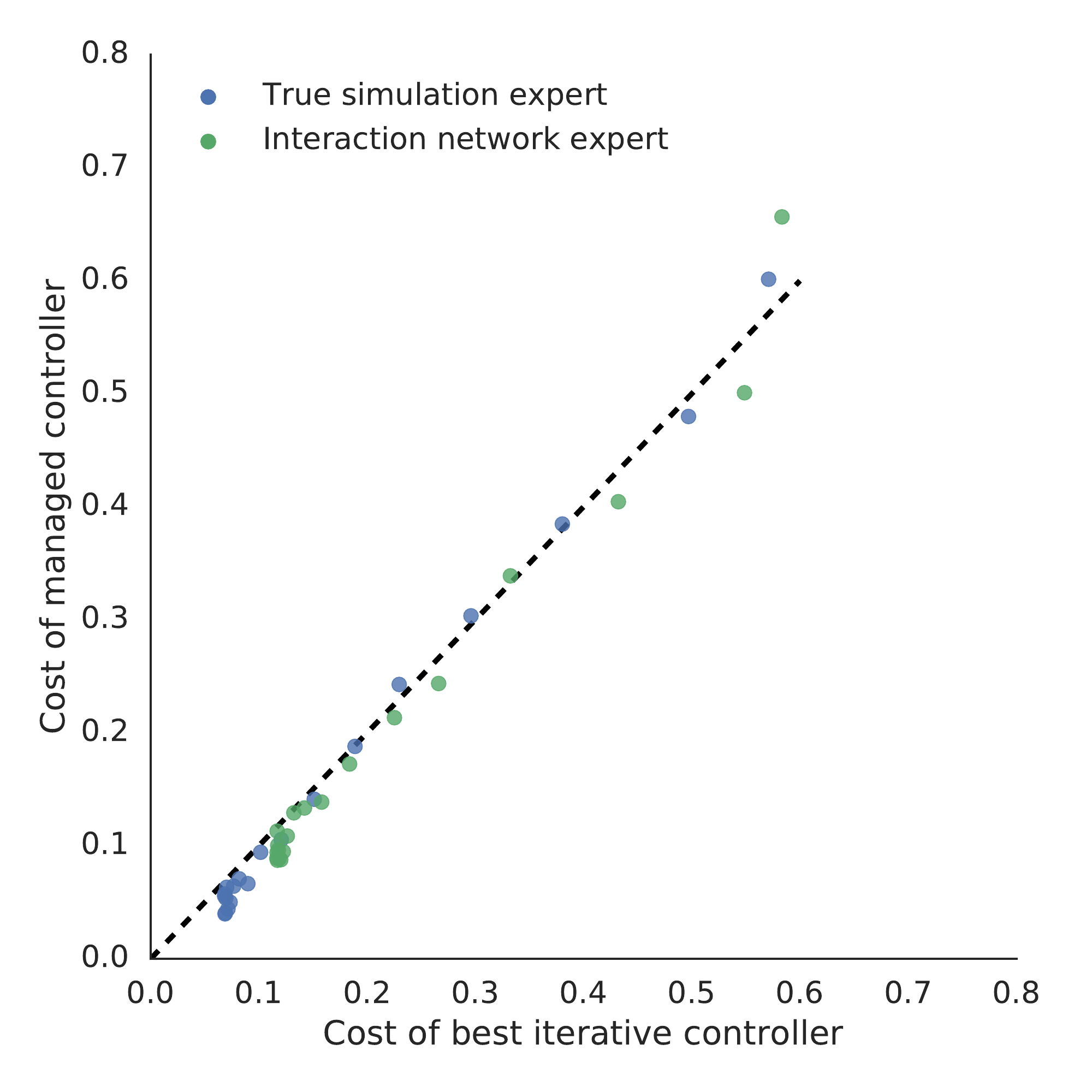}
\caption{\textbf{Cost of the best iterative controller compared to the managed controller.} Each point represents the total cost of the best iterative agent under a particular value of $\tau$ ($x$-axis) versus the total cost achieved by the metacontroller trained with the same value of $\tau$ ($y$-axis). The best iterative agent was chosen by computing the cost for all the different number of ponder steps, and then choosing the whichever number of ponder stpes yielded the lowest cost (i.e., finding the minimum of the curves in Figure~\ref{fig:reinforce}, top row). In almost all cases, the managed controller achieves a lower loss than the iterative controller: for the metacontroller with the IN expert, the cost is 11\% lower than the iterative controller on average, and for the metacontroller with the true simulation expert, it is 15\% lower on average.}
\label{fig:cost-scatter}
\end{figure}

\paragraph{Metacontroller with a single expert.} The metacontroller agent with a single expert converged more reliably than the corresponding iterative agent (see the bottom row of Figure~\ref{fig:reinforce}).
As mentioned in the previous paragraph, the iterative agent had to take more steps than actually necessary, causing it to perform less well for larger numbers of ponder steps, whereas the metacontroller agent had the flexibility of stopping when it had found a good control.
On the other hand, we found that the metacontroller agent sometimes performed too many ponder steps for large values of $\tau$ (see Figures~\ref{fig:reinforce} and \ref{fig:cost-scatter}).
We believe this is due to the entropy term ($\lambda$) added to the \textsc{Reinforce} loss.
This is because when then ponder cost is very high, the optimal thing to do is to behave deterministically and always execute (never ponder); however, the entropy term encouraged the policy to be nondeterministic.
We plan to explore different training regimes in future work to alleviate this problem, for example by annealing the entropy term to zero over the course of training.

\paragraph{Metacontroller with multiple experts.} The metacontroller agent with multiple experts was somewhat more difficult to train, especially for high ponder cost of the interaction network expert.
For example, note how the proportion of steps using the MLP expert does not decrease monotonically in Figure~\ref{fig:reinforce3} (right) with increasing cost for the MLP expert.
We believe this is also an unexpected result of using the entropy term: in all of these cases, the optimal thing to do actually is to rely on the MLP expert 100\% of the time, yet the entropy term encourages the policy to be non-deterministic.
Future work will explore these difficulties further by using experts that complement each other better (i.e., so there is not one that is wholly better than the other).

\paragraph{Experts.} The experts themselves always converged quickly and reliably, and trained much faster than the rest of the network.

\begin{table}
\centering
{\footnotesize
\begin{tabular}{cccccccc}
\toprule
& & True sim. & \multicolumn{3}{c}{MLP} & \multicolumn{2}{c}{IN}\\
\cmidrule(r){3-3}
\cmidrule(l){4-6}
\cmidrule(l){7-8}
      Dataset &  \# Ponder Steps & $\alpha_c$ & $\alpha_{c}$ & $\alpha_{E_{\rm IN}}$ & $\alpha_{E_{\rm MLP}}$ & $\alpha_{c}$ & $\alpha_{E_{\rm IN}}$ \\
\midrule
    one planet &               0 &      1e-03 &             1e-03 &                      3e-03 &                       5e-04 &            1e-03 &                     1e-03 \\
    one planet &               1 &      1e-03 &             1e-03 &                      3e-03 &                       1e-03 &            1e-03 &                     1e-03 \\
    one planet &               2 &      1e-03 &             1e-03 &                      3e-03 &                       5e-04 &            1e-03 &                     1e-03 \\
    one planet &               3 &      1e-03 &             1e-03 &                      3e-03 &                       5e-04 &            1e-03 &                     1e-03 \\
    one planet &               4 &      1e-03 &             1e-03 &                      3e-03 &                       1e-03 &            1e-03 &                     1e-03 \\
    one planet &               5 &      1e-03 &             1e-03 &                      3e-03 &                       5e-04 &            5e-04 &                     1e-03 \\
    one planet &               6 &      1e-03 &             1e-03 &                      3e-03 &                       5e-04 &            1e-03 &                     1e-03 \\
    one planet &               7 &      1e-03 &             1e-03 &                      3e-03 &                       5e-04 &            1e-03 &                     1e-03 \\
    one planet &               8 &      1e-03 &             1e-03 &                      3e-03 &                       1e-03 &            1e-03 &                     1e-03 \\
    one planet &               9 &      5e-04 &             1e-03 &                      3e-03 &                       5e-04 &            5e-04 &                     1e-03 \\
    one planet &              10 &      1e-03 &             1e-03 &                      3e-03 &                       5e-04 &            1e-03 &                     1e-03 \\
\midrule
   two planets &               0 &      1e-03 &             1e-03 &                      3e-03 &                       1e-03 &            3e-03 &                     3e-03 \\
   two planets &               1 &      1e-03 &             1e-03 &                      3e-03 &                       5e-04 &            1e-03 &                     1e-03 \\
   two planets &               2 &      1e-03 &             1e-03 &                      3e-03 &                       5e-04 &            1e-03 &                     1e-03 \\
   two planets &               3 &      1e-03 &             1e-03 &                      3e-03 &                       5e-04 &            1e-03 &                     1e-03 \\
   two planets &               4 &      1e-03 &             1e-03 &                      3e-03 &                       1e-03 &            1e-03 &                     1e-03 \\
   two planets &               5 &      1e-03 &             1e-03 &                      1e-03 &                       1e-03 &            1e-03 &                     1e-03 \\
   two planets &               6 &      1e-03 &             1e-03 &                      3e-03 &                       1e-03 &            1e-03 &                     1e-03 \\
   two planets &               7 &      5e-04 &             1e-03 &                      3e-03 &                       5e-04 &            5e-04 &                     1e-03 \\
   two planets &               8 &      1e-03 &             1e-03 &                      3e-03 &                       5e-04 &            5e-04 &                     1e-03 \\
   two planets &               9 &      1e-03 &             1e-03 &                      3e-03 &                       5e-04 &            3e-03 &                     3e-03 \\
   two planets &              10 &      5e-04 &             1e-03 &                      3e-03 &                       1e-03 &            5e-04 &                     1e-03 \\
\midrule
 three planets &               0 &      1e-03 &             1e-03 &                      3e-03 &                       1e-03 &            1e-03 &                     3e-03 \\
 three planets &               1 &      1e-03 &             1e-03 &                      3e-03 &                       1e-03 &            1e-03 &                     1e-03 \\
 three planets &               2 &      1e-03 &             5e-04 &                      3e-03 &                       1e-03 &            1e-03 &                     1e-03 \\
 three planets &               3 &      1e-03 &             1e-03 &                      1e-03 &                       5e-04 &            1e-03 &                     1e-03 \\
 three planets &               4 &      1e-03 &             1e-03 &                      3e-03 &                       5e-04 &            1e-03 &                     1e-03 \\
 three planets &               5 &      1e-03 &             1e-03 &                      1e-03 &                       5e-04 &            5e-04 &                     1e-03 \\
 three planets &               6 &      1e-03 &             5e-04 &                      3e-03 &                       5e-04 &            1e-03 &                     1e-03 \\
 three planets &               7 &      1e-03 &             1e-03 &                      3e-03 &                       1e-03 &            1e-03 &                     1e-03 \\
 three planets &               8 &      1e-03 &             1e-03 &                      3e-03 &                       1e-03 &            5e-04 &                     1e-03 \\
 three planets &               9 &      1e-03 &             1e-03 &                      3e-03 &                       5e-04 &            1e-03 &                     1e-03 \\
 three planets &              10 &      1e-03 &             5e-04 &                      3e-03 &                       1e-03 &            1e-03 &                     1e-03 \\
\midrule
  four planets &               0 &      1e-03 &             5e-04 &                      3e-03 &                       5e-04 &            1e-03 &                     1e-03 \\
  four planets &               1 &      1e-03 &             5e-04 &                      3e-03 &                       1e-03 &            1e-03 &                     1e-03 \\
  four planets &               2 &      1e-03 &             5e-04 &                      3e-03 &                       1e-03 &            1e-03 &                     1e-03 \\
  four planets &               3 &      1e-03 &             1e-03 &                      3e-03 &                       5e-04 &            1e-03 &                     1e-03 \\
  four planets &               4 &      1e-03 &             5e-04 &                      3e-03 &                       1e-03 &            1e-03 &                     1e-03 \\
  four planets &               5 &      1e-03 &             1e-03 &                      3e-03 &                       1e-03 &            1e-03 &                     1e-03 \\
  four planets &               6 &      1e-03 &             1e-03 &                      3e-03 &                       1e-03 &            1e-03 &                     1e-03 \\
  four planets &               7 &      5e-04 &             1e-03 &                      1e-03 &                       1e-03 &            1e-03 &                     1e-03 \\
  four planets &               8 &      5e-04 &             1e-03 &                      3e-03 &                       1e-03 &            1e-03 &                     1e-03 \\
  four planets &               9 &      1e-03 &             1e-03 &                      3e-03 &                       1e-03 &            5e-04 &                     1e-03 \\
  four planets &              10 &      1e-03 &             1e-03 &                      3e-03 &                       1e-03 &            5e-04 &                     1e-03 \\
\midrule
  five planets &               0 &      1e-03 &             1e-03 &                      3e-03 &                       5e-04 &            1e-03 &                     3e-03 \\
  five planets &               1 &      1e-03 &             1e-03 &                      3e-03 &                       5e-04 &            1e-03 &                     1e-03 \\
  five planets &               2 &      5e-04 &             1e-03 &                      3e-03 &                       5e-04 &            1e-03 &                     1e-03 \\
  five planets &               3 &      1e-03 &             1e-03 &                      3e-03 &                       1e-03 &            1e-03 &                     1e-03 \\
  five planets &               4 &      5e-04 &             1e-03 &                      3e-03 &                       5e-04 &            1e-03 &                     1e-03 \\
  five planets &               5 &      1e-03 &             5e-04 &                      3e-03 &                       1e-03 &            1e-03 &                     1e-03 \\
  five planets &               6 &      1e-03 &             1e-03 &                      3e-03 &                       1e-03 &            1e-03 &                     1e-03 \\
  five planets &               7 &      1e-03 &             1e-03 &                      3e-03 &                       1e-03 &            1e-03 &                     3e-03 \\
  five planets &               8 &      5e-04 &             1e-03 &                      3e-03 &                       1e-03 &            1e-03 &                     3e-03 \\
  five planets &               9 &      1e-03 &             1e-03 &                      3e-03 &                       1e-03 &            1e-03 &                     1e-03 \\
  five planets &              10 &      1e-03 &             1e-03 &                      3e-03 &                       5e-04 &            1e-03 &                     1e-03 \\
\bottomrule

\end{tabular}}
\caption{Hyperparameter values for the \emph{iterative} controller. $\alpha_c$ refers to the learning rate for the controller and memory, while $\alpha_{E_\mathrm{IN}}$ refers to the learning rate for the IN expert, and $\alpha_{E_\mathrm{MLP}}$ refers to the learning rate for the MLP expert.}
\label{tbl:iterative}
\end{table}

\begin{table}
\centering
{\footnotesize
\begin{tabular}{cccccccccc}
\toprule
& \multicolumn{2}{c}{True sim.} & \multicolumn{4}{c}{MLP} & \multicolumn{3}{c}{IN}\\
\cmidrule(r){2-3}
\cmidrule(r){4-7}
\cmidrule(r){8-10}
  $\tau$ & $\alpha_c$ & $\alpha_{m}$ & $\alpha_{c}$ & $\alpha_{m}$ &  $\alpha_{E_{\rm IN}}$ & $\alpha_{E_{\rm MLP}}$ & $\alpha_{c}$ & $\alpha_{m}$ &  $\alpha_{E_{\rm IN}}$ \\
\midrule
 0.00000 &      5e-04 &        5e-04 &             5e-04 &             1e-03 &                      3e-03 &                       1e-03 &            5e-04 &            1e-04 &                     1e-03 \\
 0.00004 &      1e-03 &        1e-04 &             1e-03 &             5e-05 &                      3e-03 &                       5e-04 &            1e-03 &            1e-03 &                     1e-03 \\
 0.00006 &      5e-04 &        5e-05 &             1e-03 &             5e-04 &                      3e-03 &                       1e-03 &            5e-04 &            5e-05 &                     1e-03 \\
 0.00011 &      1e-03 &        1e-04 &             1e-03 &             1e-04 &                      3e-03 &                       1e-03 &            5e-04 &            5e-04 &                     1e-03 \\
 0.00017 &      5e-04 &        1e-04 &             1e-03 &             1e-03 &                      3e-03 &                       1e-03 &            1e-03 &            5e-05 &                     1e-03 \\
 0.00028 &      1e-03 &        1e-03 &             1e-03 &             1e-03 &                      3e-03 &                       1e-03 &            5e-04 &            5e-05 &                     1e-03 \\
 0.00045 &      1e-03 &        1e-03 &             5e-04 &             1e-04 &                      3e-03 &                       1e-03 &            1e-03 &            5e-05 &                     1e-03 \\
 0.00073 &      1e-03 &        1e-04 &             1e-03 &             1e-04 &                      3e-03 &                       1e-03 &            1e-03 &            5e-05 &                     1e-03 \\
 0.00119 &      1e-03 &        5e-05 &             1e-03 &             1e-04 &                      5e-04 &                       1e-03 &            5e-04 &            5e-04 &                     1e-03 \\
 0.00193 &      1e-03 &        5e-05 &             1e-03 &             5e-05 &                      3e-03 &                       5e-04 &            1e-03 &            5e-05 &                     1e-03 \\
 0.00314 &      1e-03 &        1e-04 &             1e-03 &             1e-04 &                      3e-03 &                       5e-04 &            1e-03 &            1e-04 &                     1e-03 \\
 0.00510 &      1e-03 &        5e-05 &             1e-03 &             5e-05 &                      3e-03 &                       1e-03 &            1e-03 &            5e-05 &                     1e-03 \\
 0.00828 &      1e-03 &        5e-04 &             1e-03 &             5e-04 &                      3e-03 &                       5e-04 &            1e-03 &            1e-03 &                     1e-03 \\
 0.01344 &      1e-03 &        5e-05 &             1e-03 &             5e-05 &                      3e-03 &                       5e-04 &            5e-04 &            5e-05 &                     1e-03 \\
 0.02182 &      1e-03 &        1e-04 &             1e-03 &             1e-04 &                      3e-03 &                       5e-04 &            1e-03 &            1e-04 &                     1e-03 \\
 0.03543 &      1e-03 &        1e-04 &             1e-03 &             1e-04 &                      3e-03 &                       1e-03 &            1e-03 &            1e-04 &                     1e-03 \\
 0.05754 &      1e-03 &        5e-04 &             1e-03 &             5e-04 &                      3e-03 &                       5e-04 &            1e-03 &            1e-04 &                     1e-03 \\
 0.09343 &      1e-03 &        5e-05 &             1e-03 &             5e-05 &                      3e-03 &                       1e-03 &            1e-03 &            1e-04 &                     1e-03 \\
 0.15171 &      1e-03 &        1e-04 &             1e-03 &             5e-04 &                      3e-03 &                       5e-04 &            1e-03 &            1e-04 &                     1e-03 \\
 0.24634 &      1e-03 &        5e-05 &             1e-03 &             1e-03 &                      3e-03 &                       1e-03 &            1e-03 &            1e-03 &                     1e-03 \\
 0.40000 &      1e-03 &        1e-03 &             1e-03 &             1e-03 &                      3e-03 &                       5e-04 &            1e-03 &            1e-03 &                     1e-03 \\
\bottomrule

\end{tabular}}
\caption{Hyperparameter values for the metacontroller with a single expert. $\tau$ refers to the ponder cost, $\alpha_c$ refers to the learning rate for the controller and memory, $\alpha_m$ refers to the learning rate for the manager, $\alpha_{E_\mathrm{IN}}$ refers to the learning rate for the IN expert, and $\alpha_{E_\mathrm{MLP}}$ refers to the learning rate for the MLP expert.}
\label{tbl:meta-one-expert}
\end{table}

\begin{table}
\centering
{\footnotesize
\begin{tabular}{cccccc}
\toprule
& & \multicolumn{4}{c}{IN + MLP}\\
\cmidrule(r){3-6}
$\tau_{\rm IN}$ &  $\tau_{\rm MLP}$ & $\alpha_{c}$ & $\alpha_{m}$ & $\alpha_{E_{\rm IN}}$ & $\alpha_{E_{\rm MLP}}$ \\
\midrule
  0.00000 &   0.00000 &        1e-03 &        5e-05 &                 1e-03 &                  1e-03 \\
  0.00000 &   0.00121 &        1e-03 &        5e-04 &                 1e-03 &                  1e-03 \\
  0.00000 &   0.00663 &        1e-03 &        1e-03 &                 1e-03 &                  1e-03 \\
  0.00000 &   0.03641 &        1e-03 &        5e-05 &                 1e-03 &                  1e-03 \\
  0.00000 &   0.20000 &        1e-03 &        5e-05 &                 1e-03 &                  1e-03 \\
  0.00000 &   0.30000 &        5e-04 &        1e-04 &                 1e-03 &                  1e-03 \\
  0.00000 &   0.40000 &        5e-04 &        5e-05 &                 1e-03 &                  1e-03 \\
\midrule
  0.00121 &   0.00000 &        1e-03 &        1e-04 &                 1e-03 &                  1e-03 \\
  0.00121 &   0.00121 &        1e-03 &        5e-05 &                 1e-03 &                  1e-03 \\
  0.00121 &   0.00663 &        1e-03 &        1e-03 &                 1e-03 &                  1e-03 \\
  0.00121 &   0.03641 &        1e-03 &        1e-04 &                 1e-03 &                  1e-03 \\
  0.00121 &   0.20000 &        1e-03 &        5e-04 &                 1e-03 &                  1e-03 \\
  0.00121 &   0.30000 &        5e-04 &        5e-05 &                 1e-03 &                  1e-03 \\
  0.00121 &   0.40000 &        1e-03 &        1e-04 &                 1e-03 &                  1e-03 \\
\midrule
  0.00663 &   0.00000 &        1e-03 &        1e-03 &                 1e-03 &                  1e-03 \\
  0.00663 &   0.00121 &        5e-04 &        5e-05 &                 1e-03 &                  1e-03 \\
  0.00663 &   0.00663 &        5e-04 &        1e-04 &                 1e-03 &                  1e-03 \\
  0.00663 &   0.03641 &        1e-03 &        1e-04 &                 1e-03 &                  1e-03 \\
  0.00663 &   0.20000 &        5e-04 &        5e-04 &                 1e-03 &                  1e-03 \\
  0.00663 &   0.30000 &        5e-04 &        1e-03 &                 1e-03 &                  1e-03 \\
  0.00663 &   0.40000 &        5e-04 &        1e-04 &                 1e-03 &                  1e-03 \\
\midrule
  0.03641 &   0.00000 &        1e-03 &        5e-04 &                 1e-03 &                  1e-03 \\
  0.03641 &   0.00121 &        1e-03 &        5e-04 &                 1e-03 &                  1e-03 \\
  0.03641 &   0.00663 &        1e-03 &        1e-03 &                 1e-03 &                  1e-03 \\
  0.03641 &   0.03641 &        1e-03 &        5e-04 &                 1e-03 &                  1e-03 \\
  0.03641 &   0.20000 &        1e-03 &        1e-04 &                 1e-03 &                  1e-03 \\
  0.03641 &   0.30000 &        1e-03 &        5e-05 &                 1e-03 &                  1e-03 \\
  0.03641 &   0.40000 &        1e-03 &        1e-04 &                 1e-03 &                  1e-03 \\
\midrule
  0.20000 &   0.00000 &        1e-03 &        5e-04 &                 1e-03 &                  1e-03 \\
  0.20000 &   0.00121 &        1e-03 &        5e-04 &                 1e-03 &                  1e-03 \\
  0.20000 &   0.00663 &        1e-03 &        5e-04 &                 1e-03 &                  1e-03 \\
  0.20000 &   0.03641 &        1e-03 &        1e-04 &                 1e-03 &                  1e-03 \\
  0.20000 &   0.20000 &        5e-04 &        1e-03 &                 1e-03 &                  1e-03 \\
  0.20000 &   0.30000 &        1e-03 &        5e-05 &                 1e-03 &                  1e-03 \\
  0.20000 &   0.40000 &        1e-03 &        5e-04 &                 1e-03 &                  1e-03 \\
\midrule
  0.30000 &   0.00000 &        5e-04 &        1e-04 &                 1e-03 &                  1e-03 \\
  0.30000 &   0.00121 &        5e-04 &        1e-03 &                 1e-03 &                  1e-03 \\
  0.30000 &   0.00663 &        1e-03 &        1e-03 &                 1e-03 &                  1e-03 \\
  0.30000 &   0.03641 &        1e-03 &        5e-04 &                 1e-03 &                  1e-03 \\
  0.30000 &   0.20000 &        1e-03 &        1e-03 &                 1e-03 &                  1e-03 \\
  0.30000 &   0.30000 &        1e-03 &        1e-04 &                 1e-03 &                  1e-03 \\
  0.30000 &   0.40000 &        1e-03 &        5e-05 &                 1e-03 &                  1e-03 \\
\midrule
  0.40000 &   0.00000 &        1e-03 &        1e-03 &                 1e-03 &                  1e-03 \\
  0.40000 &   0.00121 &        5e-04 &        1e-03 &                 1e-03 &                  1e-03 \\
  0.40000 &   0.00663 &        1e-03 &        5e-04 &                 1e-03 &                  1e-03 \\
  0.40000 &   0.03641 &        5e-04 &        1e-04 &                 1e-03 &                  1e-03 \\
  0.40000 &   0.20000 &        1e-03 &        1e-03 &                 1e-03 &                  1e-03 \\
  0.40000 &   0.30000 &        5e-04 &        1e-03 &                 1e-03 &                  1e-03 \\
  0.40000 &   0.40000 &        5e-04 &        5e-04 &                 1e-03 &                  1e-03 \\
\bottomrule
\end{tabular}}
\caption{Hyperparameter values for the metacontroller with two experts. $\tau_{\rm IN}$ refers to the ponder cost for the interaction network expert, $\tau_{\rm MLP}$ refers to the ponder cost for the MLP expert, $\alpha_c$ refers to the learning rate for the controller and memory, $\alpha_m$ refers to the learning rate for the manager, $\alpha_{E_\mathrm{IN}}$ refers to the learning rate for the IN expert, and $\alpha_{E_\mathrm{MLP}}$  refers to the learning rate for the MLP expert.}
\label{tbl:meta-two-experts}
\end{table}

{\footnotesize
\setlength{\bibsep}{2pt}
\bibliographyappendix{references}
\bibliographystyleappendix{iclr2017_conference}}

\end{document}